\documentclass{article}

\usepackage{xcolor}
\usepackage{color}
\usepackage{gensymb}
\usepackage{hyperref}
\usepackage{graphicx}
\usepackage{booktabs}
\usepackage{tabu}

\usepackage{mathtools}
\usepackage{xfrac}

\definecolor{unknownspeciescolor}{HTML}{b02000}

\newcommand{\etal}{\textit{et al}. }

\newcommand{\eg}{\textit{e}.\textit{g}.}

\begin{document}

\bibliographystyle{plain}

\title{CatchMonitor: a machine learning system for automated fish discard quantification}

\author{%
Geoff French, Michal Mackiewicz, Mark Fisher, \\
Helen Holah, Rebecca Lamb
}

\maketitle

\abstract{We report on the continued development of CatchMonitor, resulting in a prototype computer vision system designed to automatically quantify discarded fish from video footage collected from Remote Electronic Monitoring (REM) systems on fishing trawlers.
The analysis of trawler surveillance footage is a challenging problem due to the real-world conditions on board fishing vessels.
Building on our prior work we improve the accuracy of species identification through the application of semi-supervised learning.
We utilise a simple and robust object tracking approach, upon which we build our prototype discard quantification system.
Finally we analyse the variability of manual discard quantification performed by multiple expert human analysts, using it as a benchmark against which we compare the performance of our system.}

\section{Introduction}
\label{sec:introduction}


Discard estimates have traditionally been obtained by at-sea observer sampling.
This costly and human capital intensive process limits coverage levels and the availability of discard data, leading to uncertainties in discard estimates and potentially biased data~\cite{Benoit:AtSeaObserverSurveys}.
In recent years the use of Remote Electronic Monitoring (REM) systems on-board fishing vessels has gained traction.
There have been many pilot studies using REM globally~\cite{VanHelmond:ElecMonFisheries}, including a catch quota management scheme (CQMS) in the Scottish demersal fishing fleet that captured video records of the conveyor belts on which fish are processed.
Marine Scotland Science analysts obtained per-species counts and sizes of discards by manually reviewing a sample of each vessel's video record when it returned to port~\cite{Needle:REM}.

Manually reviewing video footage for discard quantification proved to be a laborious and time consuming task, motivating the development of an automated approach.
In this paper, we report on the continued development of CatchMonitor, first reported in~\cite{French:SmartfishA}.

We review a body of related work and the analysis system on which this work is based in Section \ref{sec:background}.
We discuss prior work in automated analysis of fishing data and work that underpins the computer vision components of our system.

\section{Background}
\label{sec:background}


In French \etal's work~\cite{French:SmartfishA}, we extracted still images from video footage to form the training data.
These images were manually annotated with polygonal outlines to mark each fish and identify their species, utilizing a custom-developed web-based annotation tool\footnote{Available at \url{https://github.com/Britefury/django-labeller}}.
Given that fish are randomly oriented and frequently overlap, we chose to use instance segmentation to isolate each individual.
We adopted the robust and elegant Mask R-CNN~\cite{He:MaskRCNN} algorithm for instance segmentation.
A ResNet-50~\cite{He:ResNet}-based image classifier was subsequently employed to predict the species of each isolated fish.

A succinct recap of object detection and instance segmentation - central to our system - will be provided in Section~\ref{backg:det}, eschewing detailed explanation as discussed in~\cite{French:SmartfishA}.

In~\cite{French:SmartfishA} we identified two main avenues for future work; improving the accuracy of our species classifier and developing a discard estimation system that analyses video footage.
In this work we address both these concerns.
We improved the accuracy of our species classifier using semi-supervised learning.
Quantifying discarded fish necessitates the processing of complete videos, requiring that we track each detected fish and classify those that are discarded.
We therefore review relevant prior work in the areas of semi-supervised learning and object tracking in Sections~\ref{backg:semisup} and \ref{backg:objtrack} respectively.





\subsection{Object detection and instance segmentation}
\label{backg:det}

Here we briefly cover the instance segmentation approach that we adopted in~\cite{French:SmartfishA} and continue to utilize in our system.

The object detection problem concerns finding objects of interest in an image and predicting a bounding box and class for each one.
While a variety of classic computer vision and deep learning based approaches have been proposed, here we focus on Faster R-CNN~\cite{Ren:FasterRCNN}.
Briefly, a convolutional Region Proposal Network (RPN) identifies potential objects of interest within an image by predicting \emph{objectness} scores and a bounding box refinements for anchor boxes that are arranged in a grid over the image.
A rectangular crop is extracted from a high level feature image produced by the network backbone -- usually the convolutional layers of an ImageNet classifier such as ResNet-50~\cite{He:ResNet} -- and passed to an RCNN head that predicts the class of the identified object and predicts further bounding box refinements to improve the quality of the size and position of the predicted box.
Faster R-CNN has proved to be an effective object detection approach.

Instance segmentation concerns the problem of predicting the silhouette of each object in an image.
The Mask R-CNN~\cite{He:MaskRCNN} algorithm extends a Faster R-CNN object detector, predicting a fixed resolution -- usually $28\times28$ pixels -- mask for each detected object.
The mask is scaled to cover the objects predicted bounding box, resulting in a list of identified objects alongside masks identifying the image regions that they occupy.


\subsection{Semi-supervised classification}
\label{backg:semisup}


The rise of strongly performing deep neural networks in computer vision has come at the cost of the laborious and time-intensive process of manually labelling large training datasets.
This bottleneck has prompted us to explore semi-supervised learning methods as they require ground truth labels for only a small subset of the available training data, learning from the remaining unlabelled training samples in an unsupervised fashion.

We will focus on approaches based on consistency regularization; a simple yet effective method that combines supervised cross-entropy loss with a consistency loss term that encourages stability of network predictions in response to perturbations applied to unsupervised input samples.
Several models leveraging this principle have emerged, such as the $\Pi$-model by Laine \etal~\cite{Laine:Temporal}, which applies stochastic augmentation to unsupervised input samples and minimizes the variance between predicted class probabilities.
The Mean Teacher model~\cite{Tarvainen:MeanTeachers} builds on the $\Pi$-model, encouraging prediction consistency between a student and a teacher network, whose weights are an exponential moving average of the student.

Other works instead choose to vary the source of perturbation.
Miyato \etal~\cite{Miyato:VATSemiSup} utilized adversarial perturbations that aim towards the networks' estimate of the decision boundary.
More recently RandAugment~\cite{Cubuk:RandAugment}, a rich augmentation scheme that combines 14 image transformation operations provided by the Pillow~\cite{Pillow} library has gained traction, achieving impressive results in both supervised and semi-supervised settings.
It is employed alongside Cutout~\cite{Devries:Cutout} in UDA~\cite{Xie:UDA} and the more recent state-of-the-art FixMatch~\cite{Sohn:FixMatch} models.

Mixing and masking based regularization have shown effectiveness in supervised and semi-supervised settings.
MixUp~\cite{Zhang:MixUp} improved supervised classification performance using interpolated samples resulting from blending input samples and corresponding targets.
ICT~\cite{Verma:ICT} and MixMatch~\cite{Berthelot:MixMatch} combine MixUp with consistency regularization in a semi-supervised setting.
Cutout~\cite{Devries:Cutout} masks a randomly chosen rectangular region of an image to zero, improving supervised classification.
Its effectiveness in a semi-supervised setting is demonstrated in the ablation studies in UDA~\cite{Xie:UDA} and FixMatch~\cite{Sohn:FixMatch}, where it is able to match the performance of the other 14 augmentation operations in RandAugment.
CutMix~\cite{Yun:CutMix} combines Cutout and MixUp, mixing input images by pasting a randomly chosen rectangle from one image onto another.
French \etal~\cite{French:SemiSupSeg} used it to obtain state-of-the-art results in semi-supervised semantic segmentation.
CowMask~\cite{French:MilkingCowMask} achieved impressive semi-supervised classification results by replacing the rectangular mask in Cutout and CutMix with a more flexibly shaped mask.

\subsection{Object tracking}
\label{backg:objtrack}


An object tracker identifies an object in a video frame and subsequently traces its movement across ensuing frames.
Multiple object trackers extend this to track all objects of interest, re-identifying them once they re-appear after going out of view temporarily, typically caused by occlusion.
Temporary occlusions split an objects' \emph{track} into a number of \emph{tracklets}, each one associated with a contiguous period of time during which the object is visible.

Kernelized Correlation Filters~\cite{Henriques:KCF} leverage cyclic shifts and the Fourier domain to quickly train a classifier on many positive and negative patches in a cyclic sliding window fashion.
The filter -- trained on a patch centred on a detection -- is used to locate the object in subsequent frames.
This on-line training procedure does not require a pre-trained model.

The SORT~\cite{Bewley:SORT} (Simple Online and Realtime Tracking) algorithm of Bewley \etal joins per-frame detections from an object detector (see Section~\ref{backg:det}) into tracklets.
It employs a Kalman filter to model the motion of each object, predicting its location in subsequent frames, aiding the association of detections with tracklets.
The DeepSORT extension by Wojke \etal~\cite{Wojke:DeepSORT} introduces a deep network model that extracts appearance features for the purpose of re-identification, aiding in re-associating temporarily obscured objects with prior tracklets.

Bergmann's Tracktor~\cite{Bergmann:Tracktor} re-purposes the bounding box regressor of a Faster R-CNN object detector network for tracking, eliminating the need for a dedicated tracking model.
The bounding boxes from frame $t-1$ are carried forward to frame $t$ and then adjusted using the box regressor of the R-CNN head to fit the objects' new position.
The classification component determines if the detections should be retained or discarded due to going out of view.

Zhou's CenterTrack~\cite{Zhou:CenterTrack} model -- an extension of CenterNet~\cite{Zhou:CenterNet} -- trains an object detection and tracking network that ingests the current and previous frames along with a heatmap of the detection centre points in the previous frame.
It predicts a centre point heatmap for the current frame, corresponding bounding box sizes and offsets that provide a predicted motion vector.
The offset vectors are used to associate detections between frames, propagating object identity.

Karthik \etal~\cite{Karthik:UnsupMOT} present an unsupervised method for training an object re-identification model.
They apply a pre-trained object detection network frame-by-frame to detect objects of interest in video segments and apply the SORT algorithm to join them into noisy tracklets.
The tracklets are used to train a re-identification model to associate presentations of an object from different frames, while discriminating between different objects.
At inference they employ the DeepSORT approach; a CenterNet object detector and SORT are used to obtain tracklets, with the re-identification model used to combine them into complete tracks.

We note that both DeepSORT and CenterTrack require an object tracking dataset with ground truth tracks.
CenterTrack leverages successive frame pairs and their corresponding tracks for training the detection and tracking model, while DeepSORT trains the appearance extraction model using various object views from different frames.

\subsection{Automated fisheries monitoring}
\label{backg:fisheries}

Tseng \etal~\cite{Tseng:DetectingFish} developed a system for automatically counting fish in surveillance footage capturing the deck of a longline fishing vessel.
Footage of the deck of a fishing vessel presents challenging conditions due to weather and varying day and night lighting conditions.
They used a Mask R-CNN instance segmentation model to detect fish and a simple distance thresholding to avoid double counting between subsequent frames.
They were able to achieve strong performance on four classes: tuna, marlin, shark and \emph{other}.

Van Essen \etal~\cite{VanEssen:AutoDiscard} developed a discard quantification for conveyor belt footage, exploring a similar problem to CatchMonitor.
They used a YOLOv3~\cite{Redmon:YOLOv3} object detector to detect and classify fish in individual frames and a SORT~\cite{Bewley:SORT} based tracker to associate detections across frames.
Species predictions for a tracked fish are aggregated across frames using Bayes theorem.
The discarded catch from beam trawlers was gathered and manually counted to obtain accurate ground truth data.
The discards were placed on a conveyor belt with a camera and lighting rig mounted above it.
The footage was analysed both automatically and by human analysts, allowing comparison.
The training data was extended with synthetic data generated by inserting images of manually segmented fish against empty conveyor belt background images.

The discard quantification system developed by Sokolova \etal~\cite{Sokolova:Integrated} used a YOLOv5~\cite{Jocher:YOLOv5} object detection model extended with a regression output that predicts the weight of a detected fish.
While other systems have used object tracking to avoid double counting, Sokolova \etal developed a semi-linescan approach, in which small segments of subsequent frames captured by a camera above the moving conveyor belt are joined to form a continuous image.

\section{Method}
\label{sec:method}

We adopt the Mask R-CNN~\cite{He:MaskRCNN} instance segmentation and species classification approaches discussed in~\cite{French:SmartfishA}.
With the exception of some changes to the belt motion estimation algorithm -- see Appendix~\ref{app:belt_motion} -- and the annotation tool -- see Appendix~\ref{app:anno_tool} -- we leave the system presented in~\cite{French:SmartfishA} as is.
Our hyper-parameter optimization procedure -- see Section~\ref{sec:discard:hyperopt} -- resulted in us using a detection confidence threshold $\tau_d$ of 0.7 and a mask NMS threshold $\tau_n$ of 0.95.

We explore the use of semi-supervised learning to alleviate the annotation bottleneck; a common problem in practical machine learning scenarios.

\subsection{Semi-supervised learning}
\label{sec:method:semi_sup}

The labour intensive and time consuming task of annotating the training images -- performed by staff at Marine Scotland Science and CEFAS -- proved to be a bottleneck, resulting in annotations for only a subset of the training images in our dataset.
This situation was a natural fit for semi-supervised learning that uses unannotated samples to improve performance beyond that which can be achieved with traditional supervised learning.

We explored the use of semi-supervised learning for both instance segmentation and species classification.
We focused on approaches based on consistency regularization~\cite{Oliver:RealisticEval} due to its simplicity and the state of the art results achieved in semi-supervised classification~\cite{Sohn:FixMatch} and semantic segmentation~\cite{French:SemiSupSeg}.

We experimented with semi-supervised instance segmentation by adapting the approach of French \etal~\cite{French:SemiSupSeg} for Mask R-CNN.
Unfortunately these experiments did not yield positive results.
Conversely our results for species classification were far more promising, yielding improved classifier performance, in spite of the difficult nature of our species classification dataset.

\subsection{Species classification}
\label{sec:method:specclf}

As in ~\cite{French:SmartfishA} and in contrast to other work~\cite{VanEssen:AutoDiscard,Sokolova:Integrated}, we do not train our our object detection model to perform the task of classification at the same time.
Instead, we choose to extract the individual fish and horizontally align them before passing them to a separate species classification model that is a 50-layer residual network~\cite{He:ResNet} adapted and fine tuned using transfer learning.
Individuals are \emph{cut out} using ground truth outlines during training, or masks predicted by Mask-RCNN at inference time.

The list of species grew considerably since our prior work in~\cite{French:SemiSupSeg} to a total of 42 species, although 4 of these species (conger eel, brown crab, European lobster and crawfish) do not have any samples.


The pre-processing and data augmentation remain unchanged from~\cite{French:SemiSupSeg}.
Briefly, each fish was scaled to fit within the central region $196 \times 196$ region of a $224 \times 224$ image.
To counter errors in the predicted instance masks, they are dilated by 5\% of the region size (10 pixels).
The orientation of the fish is estimated using \texttt{regionprops} from Scikit-Image~\cite{DerWalt:skimage} and the image is rotated so that the fish is aligned with the horizontal axis.
Our stochastic data augmentation consists of rotations in the range $[-10^\circ,10^\circ]$, uniform scaling in the range $[0.8,1.2]$, random translations with a magnitude of 16 pixels and random horizontal and vertical flips.
This scheme was chosen manually to mimic the variance in orientation and size estimation above caused by varying mask quality.
The 16-pixel border surrounding the central $196 \times 196$ region described above was chosen in order to leave space for the mask dilation and augmentation.

Our semi-supervised training approach combines elements of FixMatch~\cite{Sohn:FixMatch} and Mean Teacher~\cite{Tarvainen:MeanTeachers}.
We minimize a combination of cross entropy loss for supervised samples and a consistency loss for unsupervised samples.
As in~\cite{Sohn:FixMatch} we use two augmentation schemes for consistency training; weak and strong.
The weak scheme, consisting of rotation, scaling, and translation, is used to generate pseudo-targets for unlabelled samples, while the strong augmentation scheme uses the richer RandAugment~\cite{Cubuk:RandAugment}.
In essence, the model is trained to yield predictions under the more challenging RandAugment that are consistent with the pseudo-targets generated under the \emph{easier} weak augmentation.
Following~\cite{Tarvainen:MeanTeachers} we use two networks; a student and a teacher.
The teacher network -- whose weights are an exponential moving average (EMA) of those of the student -- generates pseudo-targets for unlabelled samples while the student network is trained to match ground truth labels for supervised samples using cross-entropy loss and pseudo-targets for unsupervised samples using squared loss.

The effect of semi-supervised learning is evaluated in Section~\ref{sec:res_semisup}.

\subsection{Instance tracking}
\label{sec:method:tracking}

The Mask R-CNN based fish detection and instance segmentation system and species classifier discussed thus far are sufficient to quantify the fish visible in a single frame.
Counting fish within a video requires the use of an object tracker to associate presentations of an individual fish across multiple frames with one another, allowing us to count each fish only once, rather than once for each frame in which it appears.

\subsubsection{Choice of approach}

The most significant constraint on our choice of approach is the availability of training data.
A variety of approaches have been proposed for object tracking in the literature, some of which are discussed in Section~\ref{backg:objtrack}.
Some early hand-designed tracking approaches either do not require training or like KCF (kernelized correlation filters;~\cite{Henriques:KCF}) are trained on-line when the tracker is initialized with the appearance of the target object.
More recent deep learning based models are trained using object tracking datasets that consist of videos in which objects of interest have per-frame object annotations.

Creating object tracking annotations for our videos would require the development of a web-based video annotation tool as our existing tool is only designed for still images.
This would be a non-trivial task and would require a high bandwidth internet connection between our web server and our annotators in order to provide a sufficiently responsive user experience.
Furthermore, given the effort that was required to construct the segmentation and species ID training sets described previously, we did not wish impose this additional work load on our annotators.

The lack of tracking annotations limits us to approaches that do not rely on training a model on an annotated tracking dataset.
This led us to consider the SORT~\cite{Bewley:SORT} and Tracktor~\cite{Bergmann:Tracktor} approaches as both rely on object detection systems trained using datasets consisting of annotated still images rather than annotated video sequences.
Experiments with Tracktor~\cite{Bergmann:Tracktor} yielded poor results due to frequent identity switches when multiple fish are in close proximity to one another;
As a consequence we abandoned the Tracktor approach.

\subsubsection{Our tracker}

Similar to~\cite{VanEssen:AutoDiscard} our tracker is based on the SORT~\cite{Bewley:SORT} approach, but with a few modifications.
The original formulation of SORT modelled tracking targets as bounding boxes, initializing the state of a Kalman filter used to track a target with a detection from a Faster R-CNN detector.
A tracking target state consists of the object position, area and aspect ratio, along with positional and size velocities.
Estimated bounding boxes for a subsequent frame are matched to Faster R-CNN detections computing the pairwise IoU (intersection over union) overlap and applying the Hungarian algorithm~\cite{Kuhn:Hungarian}.

Our tracker does not track and estimate the size of a box using the Kalman filter.
Instead of matching boxes by IoU, we utilize the masks predicted by Mask R-CNN, matching by mask IoU score.
Any match for which the mask IoU score is lower than the threshold $\tau_m = 0.1$ (default 10\%) is discarded.
For each tracked fish we maintain an exponential moving average of the masks corresponding to its detections.
Upon being matched with a detection with corresponding mask $m_d$, the target mask $m_t$ is updated to $m_t' = \alpha_m m_t + (1 - \alpha_m)m_d$ where $\alpha_m = 0.75$ is the mask EMA factor, with a default value of 0.75.

\subsubsection{Motion model}
\label{sec:method:tracking:motion}

In pedestrian tracking scenarios -- that tracking algorithms are often evaluated on -- the apparent motion of targets arises from \emph{non-rigid} motion due to the pedestrian walking through the scene and \emph{rigid} motion resulting from the motion of the camera.
This division also makes sense in our domain.
We model the motion of individual fish as the sum of two components; the \emph{non-rigid} motion arising from a fish being manipulated by a fisher and the \emph{rigid} motion of the conveyor belt, relative to a fixed camera (see Appendix~\ref{app:belt_motion}).
Furthermore, the apparent size of pedestrians change due to moving closer to or further from the camera, where the motion of fish in the depth direction in our video footage is much more constrained, hence we do not model the change in size of our targets, as mentioned above.
Our Kalman filter models only the \emph{non-rigid} position and velocity of our targets while the belt motion estimates are used as the \emph{rigid} component.

We found that using KCF (kernelized correlation filters) to refine motion estimates improved the accuracy of our final discard counts.
KCF was originally designed to estimate the motion of an object from two image crops, with the first centred on a known position of the object and a second from the same position in a subsequent frame.
We however centre the second image crop on the object position predicted by the Kalman filter and use KCF to refine the predicted object position.

When processing a new frame, prior to matching tracking targets with detections, we estimate the position of each target in the new frame.
We first integrate the targets' non-rigid velocity $v_n$ -- obtained from the Kalman filter state -- over the time step to estimate its non-rigid offset $\Delta p_n = v_n \Delta t$.
This is summed with the rigid belt offset $\Delta p_r$ to obtain the total offset $\Delta p_t = \Delta p_r + \Delta p_n$.
We offset the bounding box in the current frame $b_0$ with the offset $\Delta p_t$ to estimate the position of the bounding box in the new frame $b_1$.
We then estimate the residual offset $\Delta p_d$ using the KCF algorithm, given image regions extracted from the current and new frame corresponding to $b_0$ and $b_1$ respectively.
We compute an observed non-rigid velocity $v_{obs} = \frac{\Delta p_n + \Delta p_d}{\Delta t}$ and update the Kalman filter state to estimate the position of the object in the new frame, given the previous estimate of the position and velocity and the observed velocity $v_{obs}$.

The original SORT~\cite{Bewley:SORT} algorithm stops tracking an object if it is absent for one or more frames due to not being detected, as `the constant velocity model is a poor predictor for the true dynamics' of their problem domain and they do not attempt object re-identification.
In contrast we allow fish to be re-acquired by the tracker after being absent for a maximum of $\eta_m = 7$ frames.
This allows fish to be missing from the list of detections for brief periods of time, which can occur due to being obscured by the actions of fishers or detection failures.

We obtained a small improvement in discard count accuracy by not updating the Kalman filter motion model for a tracked target when no detection is matched with it.
Missed detections occur due to mis-predictions by Mask R-CNN or due to the fish being temporarily obscured.
Manual observations revealed that in such cases the Newtonian motion modelled by the Kalman filter caused the predicted position of a tracked fish to move with a constant velocity from the time of its last successful detection.
The fish in our surveillance videos tend to come to a fairly abrupt stop when they are slowed by friction or collisions with other fish.
Our Kalman filter therefore applies the effect of the velocity of the fish to its position for one frame, after which the velocity in the Kalman filter state is largely damped to zero.
This yielded mildly superior performance to a position only motion model that uses only a KCF to search for the tracked object at its last known position, with a belt motion offset applied.

\subsection{Discard counter}
\label{sec:method:discard}

Our discard counter application quantifies the discarded fish in a video.
It combines the aforementioned instance segmentation, species classification and object tracking components with a final discard detection step.



\begin{figure}[t]
  \centering
  \includegraphics[width = 0.6\textwidth]{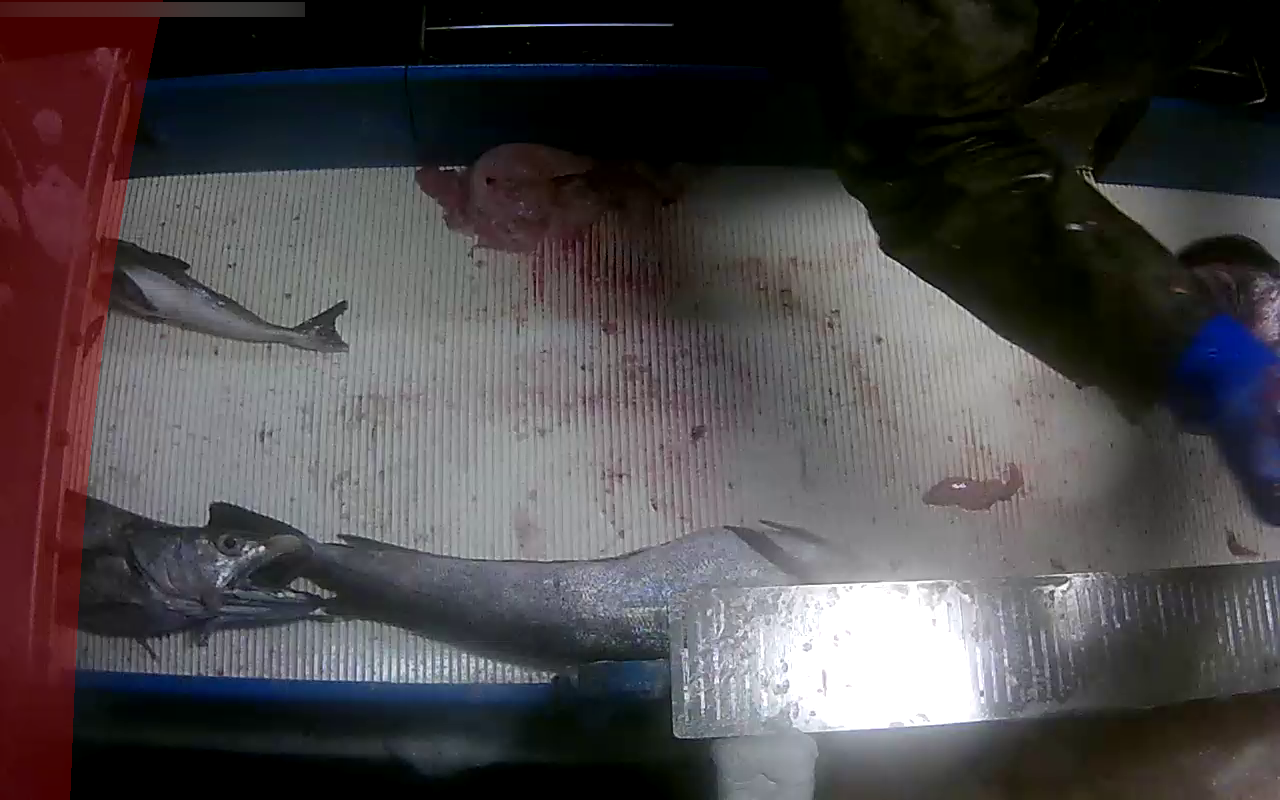}
  \caption[Discard region]{The discard region of vessel A shown on the left in red.}
\label{fig:fishdiscard:discard_region}
\end{figure}

We identify discarded fish by determining if a tracked fish crosses into the discard region; a region of the image that lies beyond the edge of the belt, covering the discard chute (see Figure~\ref{fig:fishdiscard:discard_region} for an example).
We represent the discard region as a binary mask image that we compute using the lens distortion parameters and the estimated perspective transform of the belt (see \cite{French:SmartfishA}).

As stated above, our Kalman filter and KCF based motion model (Section~\ref{sec:method:tracking:motion}) predict per-target inter-frame motion offsets that are used to move the targets' mask $m_t$.
For each target we maintain the cumulative discarded area, which we increment at each frame by the area of the intersection of $m_t$ and the discard region mask. 
When the ratio of the cumulative discarded area to the total mask area $|m_t|$ rises above a discard threshold $\tau_z = 0.333$, the fish is considered to have been discarded provided that it was detected for a minimum of $\eta_z = 6$ frames.
Tracked objects detected for less than $\eta_z$ frames that cross into the discard region could arise from false detections, so they do not contribute to the discard count.

Fish that are retained for landing at port are typically removed from the belt by the fishers before they cross into the discard region, preventing them from contributing to the discard count, as is desired.
We observed that fishers rapidly manipulate and gut fish that are to be retained, often resulting in tracking errors due to being lost by the tracker and re-detected as new targets.
Given that they do not cross into the discard region, these lost tracks do not contribute to the discard count.
The end goal of discard quantification allows us to ignore such errors as they have little effect on the final output of the system.

\begin{figure}[t]
  \centering
  \includegraphics[width = 0.95\textwidth]{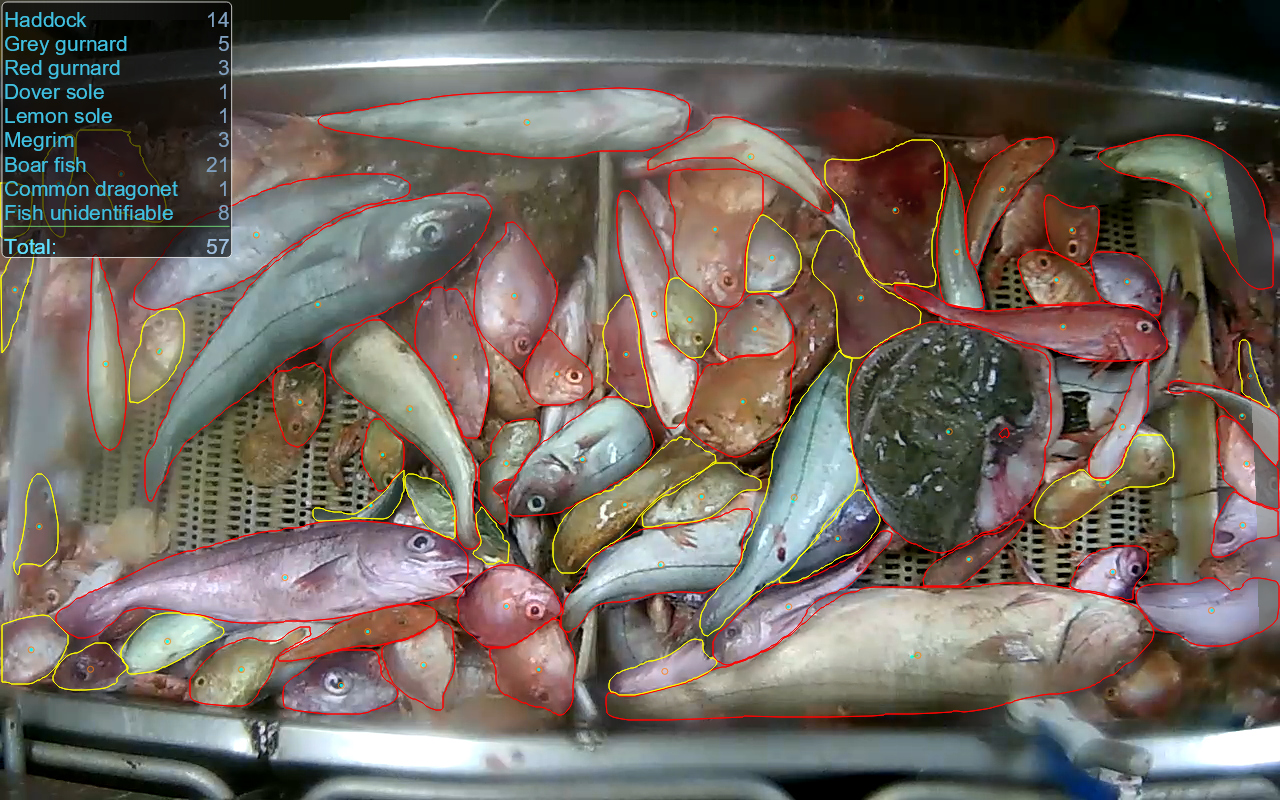}
  \caption[Discard counter video output]{A frame from the output video generated by the discard counter, applied to one of the test videos from vessel F.
  The predicted mask for each fish are used to highlight them with a partial opacity fill in a randomly chosen pastel colour.
  Fish that will ultimately be predicted to be discarded by our system are outlined in red, otherwise they are outlined in yellow.}
\label{fig:fishdiscard:discard_out_frame}
\end{figure}

We use our species classifier to predict the species of any target that is marked as a discard, averaging the species predictions arising from images extracted from multiple video frames for each fish to improve accuracy.
Using only the most recent frame in which it was visible -- as it falls into the discard chute -- would compromise accuracy due to it being partially visible and having poor definition due to motion blur.

To support species classification using multiple frames, we maintain a list of images for every tracked fish.
For each video frame we extract an aligned image of every tracked fish that was detected, following the same pre-processing procedure as used during the training of the species classifier (see Section~\ref{sec:method:specclf}).
The image is added to the list that we cap at 50 images per fish, retaining those images with the largest on screen area, so as to maximize the available visual detail.
When a fish is marked as a discard we predict the species for each image in its list and compute the weighted average of the resulting probability vectors, using the on-screen area as the weight.
A discard record consisting of the predicted species and the time of discarding relative to the start of the video is appended to a discard log.
Detections that are identified as either \emph{unidentifiable fish} or \emph{not fish} -- classes present in the training set -- must be treated specially by down-stream applications for the purpose of evaluation or final use.


The output of our discard counter application is a CSV file containing the discard log and optionally, visualization data for rendering a video depicting the fish counting process.
The visualisation outlines detected fish and highlights those that will be discarded, optionally showing the predicted species.
A cumulative discard count is displayed in the top left of the video.
Please see Figure~\ref{fig:fishdiscard:discard_out_frame} for an example.

\section{Dataset}

Our data consists of 800p HD resolution surveillance footage that was stored in MPEG-4 format.
It captures the real-world working environment and presents challenging conditions, including occlusions by personnel working at the conveyor belt and the view being obscured by spatter on the dome that covers the camera.
Fish are loaded -- often off-screen -- onto the conveyor from a fish pound or manually by the fishers.
One or more fishers standing at the belt pick up and gut the individual fish that are to be landed at port.
Once this saleable fish has been removed and processed, the conveyor belt is activated and it carries the remaining fish to be discarded, guts and detritus over the edge into a discard chute that leads to the sea.

The footage is summarised in Table \ref{tab:dataset:video}.
Footage from vessels A-D was provided by Marine Scotland Science, while footage from vessels E and F was provided by CEFAS.

\begin{table}[h!t]
  \begin{center}
  \footnotesize
  \begin{tabular}{|l|rr|rr|rr|}
  \hline
  Vessel              & \multicolumn{2}{c|}{Training}         & \multicolumn{2}{c|}{Validation}       & \multicolumn{2}{c|}{Test}             \\
                      & \# Videos             & Time          & \# Videos             & Time          & \# Videos             & Time          \\
  \hline
  
  Vessel A            & 38                     & 37:30:47     &  4                    & 4:00:02      &  5                     & 5:00:02      \\
  Vessel B            & 23                     & 22:45:42     &  5                    & 5:00:02      &  5                     & 5:00:02      \\
  Vessel C            & 26                     & 20:38:26     &  4                    & 2:58:56      &  5                     & 3:47:37      \\
  Vessel D            & 25                     & 24:14:22     &  5                    & 4:22:15      &  5                     & 4:45:50      \\
  Vessel E            & 26                     & 2:30:59      &  9                    & 1:01:42      & 13                     & 1:01:19      \\
  Vessel F            &  8                     & 2:44:13      &  8                    & 0:35:00      &  9                     & 0:42:25      \\
  
  \hline
  \end{tabular}
  
  \caption[Summary of video footage]{Summary of video footage. Running times are of the form HH:MM:SS.}
  \label{tab:dataset:video}
  
  \end{center}
  \end{table}

\section{Results: semi-supervised species classification}
\label{sec:res_semisup}

Our species classification dataset was created by extracting images of individual fish from frames that were selected for annotation.
Some were annotated by our annotators and were manually assigned a species, while others were labelled automatically using our Mask-RCNN based instance segmenter and assigned an \emph{unknown} species class.
The dataset consists of 27,897 individual fish.
A more detailed summary can be seen in Appendix~\ref{sec:spec_id_ds_summ} Table~\ref{tab:species:commercial_data}.

In order to evaluate our species classifiers we split our dataset into four folds for K-fold cross validation.
To minimize the chance of cross-contamination between our data splits we assigned a fold to each video and propagated the fold assignments to the fish according to the video they were observed in.
Some frames selected for annotation from the same video can be sufficiently near to one another that they could feature different presentations of the same individual fish.

The effects of semi-supervised learning on classifier performance are summarised in Table~\ref{tab:species:clf:acc_comm_comm}.
Its use results in an 8.3\% or 9.4\% increase in mean class accuracy when using CowMix~\cite{French:MilkingCowMask} or RandAugment~\cite{Cubuk:RandAugment} respectively.
A more detailed analysis along with confusion matrices can be found in Appendix~\ref{sec:spec_id_cm}.
In general we observe that semi-supervised learning improves the per-class recall for classes that are poorly represented in the training data.
This is a welcome observation given the high class imbalance present in our dataset.

\begin{table}[h!t]
  \begin{center}
  \scriptsize
  \begin{tabular}{ll}
  
  \hline
  Training approach               & Mean class accuracy (\%)      \\
  \hline
  Supervised                      & 44.4\%                        \\
  Semi-supervised, CowMix         & 52.7\%                        \\
  Semi-supervised, RandAugment    & 53.8\%                        \\
  \hline
  \end{tabular}
  
  \caption[Supervised and semi-supervised species ID performance]{Performance of supervised and semi-supervised species classification using four-fold cross validation.}
  \label{tab:species:clf:acc_comm_comm}
  
  \end{center}
\end{table}

\section{Results: discard counter}
\label{sec:discard}



Expert analysts at Marine Scotland Science and CEFAS manually counted discards in validation and test videos from vessels A, B, C, D, E and F (see Table~\ref{tab:dataset:video}).
The validation videos were used for hyper-parameter tuning and each video was analysed by one analyst.
The testing videos -- used for final evaluation -- were analysed by multiple analysts, allowing us to evaluate the accuracy of our system in comparison with the average of multiple analysts and in light of inter-analyst variability.
The ground truth data that we received from the expert analysts for a given video takes the form of a spreadsheet in which each row corresponds to a discarded fish, providing its species and the time at which it was annotated as a discard.

\subsection{Evaluating model predictions}
\label{sec:discard:eval}

We noticed that the set of classes used in the ground truth discard counts do not match the set of classes used in our training data.
We therefore established a mapping between the two sets of classes to enable comparison.
This mapping is presented in Appendix~\ref{sec:discard_species_map}.

The intended use case of our system -- that of quantifying discards for fisheries scientists and regulators -- suggests a comparison of the per-species predicted and ground truth counts as the most relevant metric.
We compute count accuracy as:

\begin{equation}
\frac{\min(P_c, T_c)}{\max(P_c, T_c)}
\label{eqn:count_acc}
\end{equation}

where $P_c$ and $T_c$ are the number of predicted and ground truth discards of species class $c$ respectively.
This formulation will penalise the accuracy score in cases where our system either under or over counts fish of a given species.

Given that a predicted count for a span of footage can conceal over-counting and under-counting errors that cancel one another out, we also report the precision, recall and F1 metrics computed at the level of individual fish.
The true positive counts used to compute them requires that we match predicted and ground truth discards that refer to the same individual.

\subsubsection{Matching discards}

Ideally we would match per-frame instance segmentation -- or at least bounding box -- predictions and ground truth annotations, however the manual labour required to create such annotations for every fish in our evaluation videos makes this an infeasible proposition.

We must therefore match discard records from the ground truth and predicted discard logs that provide the species and time at which each fish was discarded.
We match discards using their species and an estimate of their position along the travel direction of the conveyor belt, that we obtain by converting the discard time to a frame number and looking up the cumulative belt motion for that frame.
We represent the time to position function as $\rho(\cdot)$.


We match $M$ ground truth discards with $N$ predicted discards by applying the Hungarian algorithm~\cite{Kuhn:Hungarian} to a $M \times N$ cost matrix $A$:

\[
    A_{i,j}= 
\begin{cases}
    |\rho(s_i) - \rho(t_j)| + p(c_i, d_j)w_{max},    & \text{if } |\rho(s_i) - \rho(t_j)| \le w_{max} \\
    10^5,                                            & \text{otherwise} \\
\end{cases}
\]

$c_i$ and $d_j$ are the ground truth and predicted discard species classes respectively and $s_i$ and $t_j$ are the discard times.
To account for differences that can arise between when our algorithm and when a human analyst consider a fish to be discarded, we allow discards to match if their estimated positions are less than $w_{max}$ metres apart (with a default of 0.5), assigning a large \emph{no-match} cost of 100,000 otherwise.

A species class match penalty $p(c_i, d_j)$ specified in Tables~\ref{tab:count:count_cls_map_mss} and \ref{tab:count:count_cls_map_cefas} encourages discards of the same or compatible classes to be matched with one another.
We use a default \emph{bad match} penalty of 10 for classes that do not match.
The penalty multiplied by $w_{max}$ contributes to the cost.

The matches resulting from applying the Hungarian algorithm to $A$ are used to compute per-class precision, recall and F1 scores across either individual videos or over all of the videos from a given vessel.

\subsubsection{Extended confusion matrices}

We compute the aforementioned metrics from an extended confusion matrix that we derive from the matches obtained from the Hungarian algorithm.
The confusion matrix quantifies the predictions and mis-predictions of the system on a per-species basis and is extended with an additional row and additional column that respectively quantify false positives (predicted discards that were not matched to a ground truth discard) and false negatives (ground truth discards that were not detected).

\subsubsection{Evaluation using multiple analysts}

We assess the performance of our system using the average of the output of multiple analysts as the ground truth.
We do this by computing our performance metrics from an un-normalized extended confusion matrix $C = \frac{\sum_{o}^{S}{C_o}}{|S|}$ that is the element-wise average of the un-normalized confusion matrices $C_o$ for each analyst $o$ in the set of analysts $S$.
Each $C_o$ compares our system to the ground truth provided by analyst $o$.

We also assess the performance of our system in light of inter-analyst variability.
We quantify inter-analyst variability by computing the average performance of any given expert analyst in comparison to the other analysts in the study.
For each expert analyst $p \in S$ we compute the the mean confusion matrix $B_p$ that is the per-element mean of $D_{p,q}$ for each $q \in S, q \ne p$.
$D_{p,q}$ is an un-normalized confusion matrix that assesses the observations from analyst $p$ against those of analyst $q$.
From this we can compute performance metrics -- normalized confusion matrix, precision and recall -- for the analyst $p$.
We then compute the average of the performance by computing the mean of the performance metrics for all analysts $p \in S$.

\subsection{Hyper-parameter optimization}
\label{sec:discard:hyperopt}

The videos in our validation set were used to select optimal hyper-parameter values that were then used to obtain predictions for the videos in the test set.
We optimized the $\tau_d$, $\tau_n$, $\tau_m$, $\alpha_m$, $\eta_m$, $\tau_z$ and $\eta_z$ hyper-parameters described in earlier sections, obtaining the optimal values given for each one that we provided with their definition.

We opted for a manual optimization process in which we tuned each hyper-parameter in turn, testing a range of manually chosen values and selecting the value that resulted in the best outcome.

We developed a fitness score that blends metrics indicative of practical real-world performance and used it to guide our optimization process.
Our fitness score is a weighted average of the following metrics covering all vessels: the \emph{video count accuracy} averaged over all species with a weight of 2; the \emph{count accuracy} averaged over all species with a weight of 1; the per-species F1-score computed for each vessel, then averaged with a weight of 1; and finally the per-species F1-score computed for each video, then averaged with a weight of 1.

\begin{table}[h!t]
  \begin{center}
  \footnotesize
  \resizebox{\linewidth}{!}{%
  \begin{tabu}{llrrrrrrrrrr}
  \toprule
                      & \# Samples       &             &              & Video       &            &            &            & Vessel     &                &           &           \\
                      & in training      &  True       & Predicted    & count       & Video      & Video      & Video      & count      & Vessel         & Vessel    & Vessel    \\
              Species & data             &  count      & count        & accuracy    & precision  & recall     & F1         & accuracy   & precision      & recall    & F1        \\
  \midrule
         Anglerfish &               334 &       37.5 &       98.0 &      31.29\% &     10.99\% &    24.04\% &   11.62\% &    38.27\% &        12.59\% &     32.89\% & 18.20\% \\
         Argentines &               127 &      146.2 &      332.0 &      38.72\% &     19.45\% &    49.66\% &   24.42\% &    44.03\% &        24.90\% &     56.56\% & 34.58\% \\
                Bib &               428 &      707.0 &      699.0 &      43.34\% &     24.99\% &    46.99\% &   24.83\% &    98.87\% &        75.97\% &     75.11\% & 75.53\% \\
\rowfont{\color{unknownspeciescolor}}
       Blue whiting &                 0 &       56.3 &        0.0 &       0.00\% &         -- &     0.00\% &    0.00\% &     0.00\% &            -- &      0.00\% &  0.00\% \\
          Boar fish &              1002 &     1767.7 &     1259.0 &      32.01\% &     52.86\% &    47.18\% &   35.82\% &    71.22\% &        81.15\% &     57.80\% & 67.51\% \\
              Brill &                22 &        0.0 &       25.0 &       0.00\% &      0.00\% &        -- &    0.00\% &     0.00\% &         0.00\% &         -- &  0.00\% \\
            Catfish &                 2 &        0.7 &        1.0 &      66.67\% &      0.00\% &     0.00\% &    0.00\% &    66.67\% &         0.00\% &      0.00\% &  0.00\% \\
                Cod &               447 &        2.3 &      136.0 &       1.39\% &      0.53\% &    16.67\% &    0.81\% &     1.72\% &         0.49\% &     28.57\% &  0.96\% \\
         Common dab &                46 &       38.8 &       51.0 &      30.62\% &     15.00\% &    13.71\% &   10.28\% &    76.14\% &        11.76\% &     15.45\% & 13.36\% \\
    Common dragonet &               203 &      162.7 &      265.0 &      32.60\% &     16.65\% &    49.47\% &   19.31\% &    61.38\% &        18.99\% &     30.94\% & 23.54\% \\
         Conger eel &                 1 &        0.0 &        3.0 &       0.00\% &      0.00\% &        -- &    0.00\% &     0.00\% &         0.00\% &         -- &  0.00\% \\
\rowfont{\color{unknownspeciescolor}}
               Crab &                 0 &      412.0 &        0.0 &       0.00\% &         -- &     0.00\% &    0.00\% &     0.00\% &            -- &      0.00\% &  0.00\% \\
         Cuttlefish &                 4 &        5.0 &        3.0 &      14.58\% &      0.00\% &     0.00\% &    0.00\% &    60.00\% &         0.00\% &      0.00\% &  0.00\% \\
           Dog fish &               376 &      299.7 &      316.0 &      32.17\% &     25.00\% &    40.88\% &   21.94\% &    94.83\% &        48.63\% &     51.28\% & 49.92\% \\
         Dover sole &               360 &       25.0 &      107.0 &      14.56\% &      3.49\% &    16.00\% &    3.38\% &    23.36\% &         3.12\% &     13.33\% &  5.05\% \\
\rowfont{\color{unknownspeciescolor}}
      Great scallop &                 0 &       13.3 &        0.0 &       0.00\% &         -- &     0.00\% &    0.00\% &     0.00\% &            -- &      0.00\% &  0.00\% \\
            Gurnard &              6040 &     3724.2 &     3170.0 &      60.40\% &     66.28\% &    42.29\% &   48.06\% &    85.12\% &        66.38\% &     56.50\% & 61.05\% \\
            Haddock &              5498 &      802.2 &     1357.0 &      43.65\% &     36.21\% &    61.92\% &   37.56\% &    59.11\% &        39.76\% &     67.26\% & 49.97\% \\
               Hake &               578 &      634.7 &     1030.0 &      14.58\% &      8.01\% &    52.40\% &   10.37\% &    61.62\% &        42.94\% &     69.70\% & 53.14\% \\
            Herring &               410 &      523.2 &      638.0 &      34.82\% &     29.13\% &    58.73\% &   29.93\% &    82.00\% &        55.98\% &     68.27\% & 61.52\% \\
     Horse mackerel &               100 &       75.3 &       73.0 &      26.06\% &     25.12\% &    19.29\% &   15.44\% &    96.90\% &        43.84\% &     42.48\% & 43.15\% \\
          John Dory &                15 &        3.3 &        3.0 &      13.33\% &      0.00\% &     0.00\% &    0.00\% &    90.00\% &         0.00\% &      0.00\% &  0.00\% \\
         Lemon sole &               162 &       16.3 &       88.0 &      14.81\% &      5.88\% &    49.01\% &    7.63\% &    18.56\% &         5.30\% &     28.57\% &  8.95\% \\
\rowfont{\color{unknownspeciescolor}}
 Lesser weever fish &                 0 &        0.3 &        0.0 &       0.00\% &         -- &     0.00\% &    0.00\% &     0.00\% &            -- &      0.00\% &  0.00\% \\
               Ling &                18 &       15.2 &       60.0 &      28.61\% &      3.02\% &    11.61\% &    4.29\% &    25.28\% &         4.44\% &     17.58\% &  7.10\% \\
\rowfont{\color{unknownspeciescolor}}
            Lobster &                 0 &        6.0 &        0.0 &       0.00\% &         -- &     0.00\% &    0.00\% &     0.00\% &            -- &      0.00\% &  0.00\% \\
     Long rough dab &                34 &       17.7 &      124.0 &      14.56\% &      2.29\% &    12.71\% &    2.57\% &    14.25\% &         6.18\% &     43.40\% & 10.82\% \\
           Mackerel &                32 &       16.3 &        9.0 &      33.64\% &     19.44\% &    15.28\% &   12.31\% &    55.10\% &        29.63\% &     16.33\% & 21.05\% \\
             Megrim &               108 &       39.7 &       44.0 &      37.32\% &     26.07\% &    28.89\% &   19.88\% &    90.15\% &        31.06\% &     34.45\% & 32.67\% \\
     Norway haddock &                 4 &        5.0 &       51.0 &      16.00\% &     10.00\% &    40.00\% &    8.89\% &     9.80\% &         3.92\% &     40.00\% &  7.14\% \\
        Norway pout &               239 &      118.3 &     1790.0 &       7.43\% &      2.54\% &    54.42\% &    3.95\% &     6.61\% &         5.19\% &     78.45\% &  9.73\% \\
\rowfont{\color{unknownspeciescolor}}
            Octopus &                 0 &       15.3 &        0.0 &       0.00\% &         -- &     0.00\% &    0.00\% &     0.00\% &            -- &      0.00\% &  0.00\% \\
             Plaice &               534 &      835.3 &      763.0 &      39.28\% &     42.11\% &    56.04\% &   34.06\% &    91.34\% &        64.79\% &     59.18\% & 61.86\% \\
\rowfont{\color{unknownspeciescolor}}
           Poor cod &                 0 &       33.3 &        0.0 &       0.00\% &         -- &     0.00\% &    0.00\% &     0.00\% &            -- &      0.00\% &  0.00\% \\
\rowfont{\color{unknownspeciescolor}}
      Queen scallop &                 0 &        0.3 &        0.0 &       0.00\% &         -- &     0.00\% &    0.00\% &     0.00\% &            -- &      0.00\% &  0.00\% \\
         Red mullet &                 4 &        0.0 &        5.0 &       0.00\% &      0.00\% &        -- &    0.00\% &     0.00\% &         0.00\% &         -- &  0.00\% \\
\rowfont{\color{unknownspeciescolor}}
    Red Stone Crab  &                 0 &        2.7 &        0.0 &       0.00\% &         -- &     0.00\% &    0.00\% &     0.00\% &            -- &      0.00\% &  0.00\% \\
             Saithe &              1477 &      941.5 &     1642.0 &      30.72\% &     19.92\% &    56.10\% &   25.05\% &    57.34\% &        50.29\% &     87.71\% & 63.93\% \\
\rowfont{\color{unknownspeciescolor}}
          Scaldfish &                 0 &       19.3 &        0.0 &       0.00\% &         -- &     0.00\% &    0.00\% &     0.00\% &            -- &      0.00\% &  0.00\% \\
              Shark &                30 &        0.0 &       27.0 &       0.00\% &      0.00\% &        -- &    0.00\% &     0.00\% &         0.00\% &         -- &  0.00\% \\
          Skate/ray &               431 &      225.3 &      304.0 &      50.89\% &     36.08\% &    49.76\% &   35.02\% &    74.12\% &        48.03\% &     64.79\% & 55.16\% \\
\rowfont{\color{unknownspeciescolor}}
     Sole (generic) &                 0 &        2.0 &        0.0 &       0.00\% &         -- &     0.00\% &    0.00\% &     0.00\% &            -- &      0.00\% &  0.00\% \\
              Squid &                74 &       45.7 &       16.0 &      15.86\% &     64.81\% &    15.70\% &   18.95\% &    35.04\% &        70.83\% &     24.82\% & 36.76\% \\
\rowfont{\color{unknownspeciescolor}}
           Tuskfish &                 0 &        3.2 &        0.0 &       0.00\% &         -- &     0.00\% &    0.00\% &     0.00\% &            -- &      0.00\% &  0.00\% \\
            Whiting &               446 &      446.2 &      360.0 &      38.24\% &     26.99\% &    21.88\% &   18.65\% &    80.69\% &        43.56\% &     35.15\% & 38.91\% \\
              Witch &                 2 &        0.0 &        1.0 &       0.00\% &      0.00\% &        -- &    0.00\% &     0.00\% &         0.00\% &         -- &  0.00\% \\
     Flat (generic) &               538 &      419.0 &      565.0 &      45.36\% &     27.45\% &    42.89\% &   26.90\% &    74.16\% &        31.24\% &     42.12\% & 35.87\% \\
Fish unidentifiable &              7767 &      371.7 &     3122.0 &      16.94\% &     12.53\% &    55.89\% &   14.63\% &    11.90\% &         7.38\% &     62.02\% & 13.20\% \\
\hline
    \bf Any species &                -- &    13032.7 &    18537.0 &      60.90\% &     46.79\% &    53.55\% &   46.20\% &    70.31\% &        40.48\% &     57.58\% & 47.54\% \\
\hline
        \bf Average &                -- &    13032.7 &    18537.0 &      19.18\% &     17.58\% &    24.40\% &   10.97\% &    36.57\% &        25.79\% &     30.25\% & 20.01\% \\
  \bottomrule
  \end{tabu}
  }
  \caption[Discard count species summary for all vessels]{Discard counter performance species summary for all vessels.
  The performance of the discard counter is evaluated against the average of multiple analysts.
  Rows in dark red indicate species that were not present in the training data, thus good performance cannot be expected.
  Please see Section~\ref{sec:discard:results_test} for a description of the figures in the columns.
  }
  \label{tab:count:test_summary_all}
  
  \end{center}
\end{table}

\begin{table}[h!t]
  \begin{center}
  \footnotesize
  \resizebox{\linewidth}{!}{%
  \begin{tabu}{lrrrrrrrrr}
    \toprule
                      &            & Video        &             &            &            & Vessel     &                &             &             \\
                      &  Avg.      & count        & Video       & Video      & Video      & count      & Vessel         & Vessel      & Vessel      \\
              Species &  count     & accuracy     & precision   & recall     & F1         & accuracy   & precision      & recall      & F1           \\
    \midrule
         Anglerfish &        4.0 &      23.33\% &     41.67\% &    53.33\% &   31.11\% &    77.96\% &        42.50\% &     42.66\% & 41.74\% \\
         Argentines &      144.3 &      48.46\% &     29.90\% &    32.67\% &   25.38\% &    93.53\% &        31.88\% &     31.77\% & 31.78\% \\
       Blue whiting &       56.3 &      37.60\% &     28.99\% &    22.94\% &   16.34\% &    74.93\% &        33.58\% &     33.16\% & 32.49\% \\
            Catfish &        0.7 &      33.33\% &         -- &    66.67\% &   44.44\% &    33.33\% &            -- &     66.67\% & 44.44\% \\
                Cod &        2.0 &      30.56\% &     41.67\% &    44.44\% &   24.44\% &    35.00\% &        41.67\% &     46.67\% & 32.38\% \\
         Common dab &        4.3 &       0.00\% &         -- &     0.00\% &    0.00\% &    26.67\% &            -- &      0.00\% &  0.00\% \\
               Crab &        0.7 &      33.33\% &         -- &    66.67\% &   44.44\% &    33.33\% &            -- &     66.67\% & 44.44\% \\
         Cuttlefish &        0.3 &       0.00\% &         -- &     0.00\% &    0.00\% &     0.00\% &            -- &      0.00\% &  0.00\% \\
           Dog fish &        3.3 &      33.33\% &         -- &    40.00\% &   26.67\% &    33.33\% &            -- &     40.00\% & 26.67\% \\
            Gurnard &     2747.7 &      68.50\% &     50.15\% &    52.04\% &   47.28\% &    67.15\% &        48.74\% &     50.98\% & 47.78\% \\
            Haddock &      172.7 &      47.86\% &     30.12\% &    28.34\% &   23.83\% &    93.23\% &        33.43\% &     33.58\% & 33.44\% \\
               Hake &      634.0 &      53.59\% &     40.55\% &    42.31\% &   33.47\% &    91.93\% &        43.74\% &     43.82\% & 43.66\% \\
            Herring &      426.0 &      54.77\% &     40.06\% &    42.54\% &   35.86\% &    78.01\% &        59.97\% &     58.46\% & 58.17\% \\
     Horse mackerel &       72.0 &      34.79\% &     22.61\% &    20.20\% &   13.16\% &    74.15\% &        35.40\% &     35.49\% & 34.53\% \\
         Lemon sole &        1.0 &       0.00\% &         -- &     0.00\% &    0.00\% &     0.00\% &            -- &      0.00\% &  0.00\% \\
               Ling &       15.0 &      57.28\% &     39.72\% &    39.13\% &   33.32\% &    73.04\% &        40.87\% &     40.79\% & 39.68\% \\
            Lobster &        0.3 &       0.00\% &         -- &     0.00\% &    0.00\% &     0.00\% &            -- &      0.00\% &  0.00\% \\
     Long rough dab &       16.7 &       0.74\% &         -- &     0.00\% &    0.00\% &     3.26\% &            -- &      0.00\% &  0.00\% \\
           Mackerel &       16.0 &      39.58\% &     45.95\% &    49.36\% &   36.66\% &    95.95\% &        49.66\% &     49.77\% & 49.69\% \\
     Norway haddock &        5.0 &      69.05\% &     39.68\% &    43.06\% &   39.23\% &    69.05\% &        39.68\% &     43.06\% & 39.23\% \\
        Norway pout &      112.0 &      23.98\% &     15.97\% &    20.44\% &    9.56\% &    58.87\% &        24.89\% &     26.19\% & 23.66\% \\
            Octopus &        2.0 &      18.06\% &     16.67\% &    44.44\% &   17.86\% &    41.67\% &        16.67\% &     50.00\% & 30.00\% \\
             Plaice &        1.0 &       0.00\% &         -- &     0.00\% &    0.00\% &    41.67\% &            -- &      0.00\% &  0.00\% \\
    Red Stone Crab  &        2.7 &       8.33\% &         -- &    27.78\% &   11.11\% &    37.78\% &            -- &     35.56\% & 22.38\% \\
             Saithe &      940.0 &      47.11\% &     44.20\% &    46.25\% &   36.04\% &    75.47\% &        51.03\% &     52.08\% & 50.39\% \\
          Skate/ray &      187.0 &      50.72\% &     47.92\% &    53.62\% &   42.85\% &    75.63\% &        46.34\% &     47.76\% & 46.03\% \\
              Squid &        2.3 &      25.00\% &         -- &    22.22\% &   16.67\% &    44.44\% &            -- &     30.00\% & 23.23\% \\
           Tuskfish &        3.0 &      48.15\% &     11.11\% &    13.89\% &   10.37\% &    52.22\% &        11.67\% &     13.89\% & 11.43\% \\
            Whiting &      378.7 &      56.86\% &     41.96\% &    45.83\% &   38.06\% &    66.66\% &        42.84\% &     44.83\% & 41.93\% \\
     Flat (generic) &      248.7 &      48.11\% &     36.14\% &    37.48\% &   29.80\% &    66.93\% &        40.19\% &     41.33\% & 38.73\% \\
Fish unidentifiable &      152.7 &      43.91\% &      8.09\% &     7.70\% &    6.37\% &    37.47\% &         9.11\% &      6.90\% &  6.34\% \\
\hline
    \bf Any species &     6352.3 &      69.91\% &     46.14\% &    47.64\% &   44.05\% &    77.23\% &        45.40\% &     46.69\% & 45.20\% \\
\hline
        \bf Average &     6352.3 &      33.43\% &     31.37\% &    32.42\% &   22.40\% &    53.31\% &        34.24\% &     34.69\% & 28.85\% \\
\bottomrule
  \end{tabu}
  }
  \caption[Test inter-analyst per-species summary for all MSS vessels]{
  Inter-analyst variability species summary table for all MSS vessels, showing average metrics for the performance of each analyst vs. other analysts.
  Please see Section~\ref{sec:discard:results_test} for a description of the figures in the columns.
  }
  \label{tab:count:test_interobs_summary_all_mss}
  \end{center}
\end{table}

\begin{table}[h!t]
  \begin{center}
  \footnotesize
  \resizebox{\linewidth}{!}{%
  \begin{tabu}{lrrrrrrrrr}
    \toprule
                      &            & Video        &             &            &            & Vessel     &                &             &             \\
                      &  Avg.      & count        & Video       & Video      & Video      & count      & Vessel         & Vessel      & Vessel      \\
              Species &  count     & accuracy     & precision   & recall     & F1         & accuracy   & precision      & recall      & F1           \\
    \midrule
         Anglerfish &       33.3 &      41.87\% &     67.92\% &    53.99\% &   43.02\% &    47.76\% &        62.44\% &     53.32\% & 49.86\% \\
                Bib &      707.0 &      86.09\% &     87.25\% &    85.15\% &   84.82\% &    95.31\% &        94.39\% &     94.40\% & 94.33\% \\
          Boar fish &     1767.7 &      58.76\% &     68.52\% &    64.98\% &   57.63\% &    71.99\% &        58.76\% &     58.64\% & 56.84\% \\
                Cod &        0.3 &       0.00\% &         -- &     0.00\% &    0.00\% &     0.00\% &            -- &      0.00\% &  0.00\% \\
         Common dab &       34.3 &      48.34\% &     48.25\% &    48.89\% &   39.03\% &    78.09\% &        47.89\% &     47.04\% & 46.67\% \\
    Common dragonet &      162.7 &      74.36\% &     80.74\% &    79.12\% &   74.33\% &    90.56\% &        87.18\% &     87.08\% & 86.88\% \\
               Crab &      411.3 &      79.71\% &     81.98\% &    81.25\% &   78.77\% &    88.55\% &        83.26\% &     83.23\% & 82.86\% \\
         Cuttlefish &        4.7 &      16.67\% &     35.00\% &    42.78\% &   22.22\% &    50.28\% &        35.42\% &     43.89\% & 34.13\% \\
           Dog fish &      296.3 &      83.67\% &     83.88\% &    81.71\% &   80.58\% &    75.21\% &        80.46\% &     78.97\% & 77.80\% \\
         Dover sole &       25.0 &      52.59\% &     69.69\% &    66.84\% &   54.77\% &    92.45\% &        72.09\% &     72.18\% & 71.97\% \\
      Great scallop &       13.3 &      38.16\% &     56.44\% &    42.03\% &   34.52\% &    66.99\% &        63.62\% &     62.30\% & 59.71\% \\
            Gurnard &      941.3 &      72.74\% &     78.22\% &    77.21\% &   73.81\% &    72.44\% &        67.56\% &     66.62\% & 64.73\% \\
            Haddock &      626.7 &      77.02\% &     76.51\% &    75.19\% &   73.83\% &    80.49\% &        62.73\% &     62.09\% & 61.26\% \\
               Hake &        0.3 &       0.00\% &      0.00\% &        -- &    0.00\% &     0.00\% &         0.00\% &         -- &  0.00\% \\
            Herring &        1.0 &       0.00\% &      0.00\% &        -- &    0.00\% &     0.00\% &         0.00\% &         -- &  0.00\% \\
     Horse mackerel &        3.0 &      36.11\% &     27.78\% &    14.81\% &   13.76\% &    51.43\% &        28.33\% &     22.62\% & 22.94\% \\
          John Dory &        3.3 &      66.67\% &     81.94\% &    83.33\% &   74.60\% &    82.14\% &        80.56\% &     80.95\% & 79.85\% \\
         Lemon sole &       15.3 &      35.39\% &     64.77\% &    52.92\% &   36.40\% &    62.31\% &        69.49\% &     66.45\% & 62.69\% \\
 Lesser weever fish &        0.3 &       0.00\% &         -- &     0.00\% &    0.00\% &     0.00\% &            -- &      0.00\% &  0.00\% \\
            Lobster &        5.7 &      71.43\% &     88.89\% &    79.37\% &   71.43\% &    88.89\% &        88.89\% &     88.38\% & 88.27\% \\
     Long rough dab &        0.3 &       0.00\% &      0.00\% &        -- &    0.00\% &     0.00\% &         0.00\% &         -- &  0.00\% \\
             Megrim &       39.7 &      43.63\% &     63.03\% &    53.00\% &   41.55\% &    54.40\% &        54.32\% &     51.91\% & 47.99\% \\
        Norway pout &        4.3 &       0.00\% &      0.00\% &     0.00\% &    0.00\% &    25.00\% &         0.00\% &      0.00\% &  0.00\% \\
            Octopus &       13.3 &      60.28\% &     67.92\% &    60.79\% &   54.78\% &    84.02\% &        65.67\% &     65.47\% & 64.96\% \\
             Plaice &      834.3 &      84.55\% &     87.43\% &    86.87\% &   85.58\% &    92.34\% &        88.94\% &     88.90\% & 88.72\% \\
           Poor cod &       33.3 &      22.50\% &         -- &    32.66\% &   15.71\% &    35.08\% &            -- &     27.37\% & 17.83\% \\
      Queen scallop &        0.3 &       0.00\% &         -- &     0.00\% &    0.00\% &     0.00\% &            -- &      0.00\% &  0.00\% \\
          Scaldfish &       19.3 &      18.44\% &      5.26\% &    36.71\% &   16.16\% &    36.31\% &         6.67\% &     33.62\% & 21.56\% \\
          Skate/ray &       34.7 &      64.92\% &     72.00\% &    62.60\% &   56.33\% &    78.98\% &        77.32\% &     75.40\% & 75.22\% \\
     Sole (generic) &        2.0 &      10.42\% &     19.44\% &    20.83\% &    7.50\% &    35.00\% &        20.83\% &     23.33\% & 16.19\% \\
              Squid &       43.0 &      49.92\% &     57.31\% &    51.24\% &   43.03\% &    63.31\% &        56.67\% &     55.12\% & 52.75\% \\
            Whiting &       66.3 &      38.08\% &     54.49\% &    46.19\% &   35.63\% &    47.20\% &        59.26\% &     52.86\% & 48.28\% \\
     Flat (generic) &      163.3 &      52.56\% &     44.58\% &    42.22\% &   36.37\% &    72.02\% &        37.40\% &     37.35\% & 36.25\% \\
Fish unidentifiable &      216.3 &      41.65\% &     43.02\% &    37.22\% &   28.80\% &    62.13\% &        36.97\% &     36.75\% & 34.56\% \\
\hline
    \bf Any species &     6523.7 &      86.27\% &     74.94\% &    74.71\% &   73.96\% &    83.44\% &        68.96\% &     68.81\% & 68.22\% \\
\hline
        \bf Average &     6523.7 &      41.96\% &     56.62\% &    48.67\% &   39.26\% &    55.31\% &        55.55\% &     50.46\% & 45.44\% \\
\bottomrule
  \end{tabu}
  }
  \caption[Test inter-analyst per-species summary for all Cefas vessels]{
  Inter-analyst variability species summary table for all Cefas vessels, showing average metrics for the performance of each analyst vs. other analysts.
  Please see Section~\ref{sec:discard:results_test} for a description of the figures in the columns.
  }
  \label{tab:count:test_interobs_summary_all_cefas}
  \end{center}
\end{table}

\subsection{Results on test videos}
\label{sec:discard:results_test}

We present the per-species performance of our model in comparison to the average of our analysts on the videos across all vessels in Table~\ref{tab:count:test_summary_all}.
The \emph{true count} and \emph{predicted count} columns in Table~\ref{tab:count:test_summary_all} give the average number of ground truth discards across all analysts and the predicted number of discards respectively for each species.
The remaining metrics are discussed above in Section~\ref{sec:discard:eval} and can be grouped into either \emph{video} metrics or \emph{vessel} metrics.
The video and vessel metrics are obtained by computing the metrics separately for each video or vessel respectively and then averaging them.
At the bottom of Table~\ref{tab:count:test_summary_all} the \emph{all species} row gives performance metrics that do not consider species.
The \emph{true count} and \emph{predicted count} columns give the total number of ground truth and predicted discards.
The \emph{count accuracy} columns give count accuracies of all fish without considering species.
The precision, recall and F1 metrics are computed by first summing the true positives, false positives and false negatives across all species and then computing the metrics from these values within each video or vessel, before averaging.
The figures in the \emph{average} row are the average of the ones for each species in the rows above (excluding \emph{all species}).
They provide the average expected performance values one can expect within any class.

We would like to draw attention to our extremely strict approach for evaluating the performance of our discard counter.
The red rows in Table~\ref{tab:count:test_summary_all} correspond to species that were present in the test set but \emph{not} in the training set.
As a consequence our discard counter was unable to identify fish belonging to these species, hence receiving a zero score for all performance metrics in these rows.
These zero scores \emph{are} reflected in the \emph{average} row.
We chose this approach as it measures the real-world performance of our system on our test set and thus the expected performance in the field, with the caveat that it is not exactly fair.

We present the per-species inter-analyst performance for all MSS vessels in Table~\ref{tab:count:test_interobs_summary_all_mss} and for all Cefas vessels in Table~\ref{tab:count:test_interobs_summary_all_cefas}.
They are presented as separate tables as the set of analysts for the MSS and Cefas vessels are disjoint.

\subsection{Discussion}

Comparing the overall metrics in the \emph{any species} and \emph{average} rows of Table~\ref{tab:count:test_summary_all} with those in Tables~\ref{tab:count:test_interobs_summary_all_mss} and \ref{tab:count:test_interobs_summary_all_cefas} gives a simple summary of the performance of our discard counter.

Considering the metrics for \emph{any species} first, our discard counter achieves video and vessel count accuracies of 60.9\% and 70.31\% respectively.
They are not too far behind the count accuracies of 69.91\% and 77.23\% for MSS vessels and 86.27\% and 83.44\% for Cefas vessels achieved by human analysts.
Our system achieves video and vessel F1 scores of 46.2\% and 47.54\%; slightly ahead of the 44.05\% and 45.2\% achieved by human analysts for MSS vessels and somewhat behind the 73.96\% and 68.22\% achieved by human analysts for Cefas vessels.
These figures suggest that the detection and tracking components of our system result in overall discard estimates that are not too far away from those of human analysts.

Considering the average over all species in the \emph{average} row, we achieve video and vessel count accuracies of 19.18\% and 36.57\% and F1 scores of 10.97\% and 20.01\%.
This compares to count accuracies of 33.43\% and 53.31\% and F1 scores of 22.4\% and 28.85\% for MSS vessels and 41.96\% and 55.31\% and F1 scores of 39.26\% and 45.44\% for Cefas vessels.
These figures show a wider gap between the performance of our system and that of human analysts.

The observed difference in performance should be considered in the context of the aforementioned harsh evaluation criteria; our model receives a zero score for failing to identify species in the test set that were absent from the training set.
We also note that the disagreement between human analysts seen in Tables~\ref{tab:count:test_interobs_summary_all_mss} and \ref{tab:count:test_interobs_summary_all_cefas} would indicate that the manual species annotations in our training set used to train our models are not 100\% accurate.
We observe that discard footage manifests complex real-world situations that will confound our model.
For example many small Norway Pout are removed from the guts of larger fish that consumed them.
They are not counted by expert analysts as they were not caught by the fishers, yet they are detected and counted by our software.
This would significantly contribute to the low scores for Norway Port seen in Table~\ref{tab:count:test_summary_all}.

We should also consider that the variability in approach that human analysts take to annotating fish within an image can impact the matching of predicted discards with human annotated discards within $w_{max}$ metres. For example where there are high volumes of fish in the image it is often easier for human analysts to annotate the discards across the whole image, or partitioning the image into sections, which would result in the ground truth discard log time being well ahead of the time at which they enter the discard region, likely resulting in erroneous match failures, reducing the evaluation scores of our system.

We conclude that our prototype system shows significant promise and offer suggestions for further improving performance.
Most importantly, the size of the training set should be increased with a particular focus on improving the representation of species that are under-represented in the training set; especially those for which we have no samples at all.
Increasing the number of vessels used to gather training data would increase the diversity of visual conditions, thus improving the robustness of our models.
Furthermore we suggest that the fishers should avoid working on the last 50 to 75cm of the conveyor belt adjacent to the discard chute, reducing occlusions.
Finally, increasing the video capture frame rate to 30 frames per second would simplify the task of tracking, thereby reducing errors.

\section*{Acknowledgements}

This work was funded under the European Union Horizon 2020 SMARTFISH project, grant agreement no. 773521.

For the purpose of open access, the author has applied a CC BY public copyright licence to any Author Accepted Manuscript version arising from this submission.

\bibliography{smartfish_b}

\appendix

\section{Belt motion estimation}
\label{app:belt_motion}

The belt calibration approach described in~\cite{French:SmartfishA} combines lens distortion correction and perspective warp to rectify the conveyor belt image, simplifying motion estimation into estimating positional shifts between video frames.

In the same work, belt motion estimation was based on cross correlation of intermediate VGG-16 network features~\cite{Simonyan:VGG}, an approach quite demanding in computational terms due to the size of the neural network and the cross-correlation process applied to the numerous channels in feature images.

We have now adopted a more efficient and reliable approach inspired by panorama creation, where key-point matching between overlapping images is used to align images before compositing. Adapting this technique for motion estimation, we match key-points on the conveyor belt across successive frames, thereby estimating inter-frame motion.

The precision of the fundamental matrix used in motion estimation can be improved by constraining the model and minimizing parameter count. Applying lens distortion correction and perspective warp prior to motion estimation, we model the belt's movement with horizontal translation when initiated. A two-dimensional translation model yielded better results than a one-dimensional horizontal translation, likely due to slight inaccuracies in the belt extraction transformation.

Post-translation, the horizontal component $x$ is confined within $[-x_{max}, 0]$, where $x_{max}$ denotes maximum per-frame horizontal displacement, calculated using the maximum belt speed of $0.9ms^{-1}$, the ratio of image space to physical distance, and reciprocal of frame rate $t$. Similarly, the vertical component $y$ is limited to $[-y_{max}, y_{max}]$, with $y_{max}$ being the maximum of a vertical tolerance value of 3 pixels or $x\tan\theta_{max}$, given $\theta_{max}$ as the angular tolerance of $20\degree$.

For key-point detection and feature extraction, we used the OpenCV implementation of ORB~\cite{Rublee:ORB}. Belt motion estimation was performed from the key-points using the RANSAC~\cite{Fischler:RANSAC} algorithm, with the Scikit-Image~\cite{DerWalt:skimage} implementation being preferred due to its flexible API, permitting the usage of our constrained translation model.

Key-point and feature extraction were tested on both the original and post-extraction images. While the former method replaced a costly image mapping step with a less demanding operation of applying lens distortion correction to around 1,000 key-point coordinates, this could cause the detection algorithm to concentrate on irrelevant image sections, leading to poor estimates. Performing detection on post-extraction images prevented this issue, rendering it our chosen method.

\section{Annotation tool}
\label{app:anno_tool}

The annotation tool described in~\cite{French:SmartfishA} provides functionality for drawing and editing polygonal annotations.
The polygonal editing tools required the user to add vertices one at a time using mouse clicks.
While usable, this is a labour intensive work-flow.
The tool has since been extended with a brush tool that allows the user to paint regions.
The user may also choose to add or remove sections of a label via the use of boolean operations, provided by the polybooljs~\cite{polybooljs} library.

\begin{figure}
  \centering
  \includegraphics[width = 0.8\textwidth]{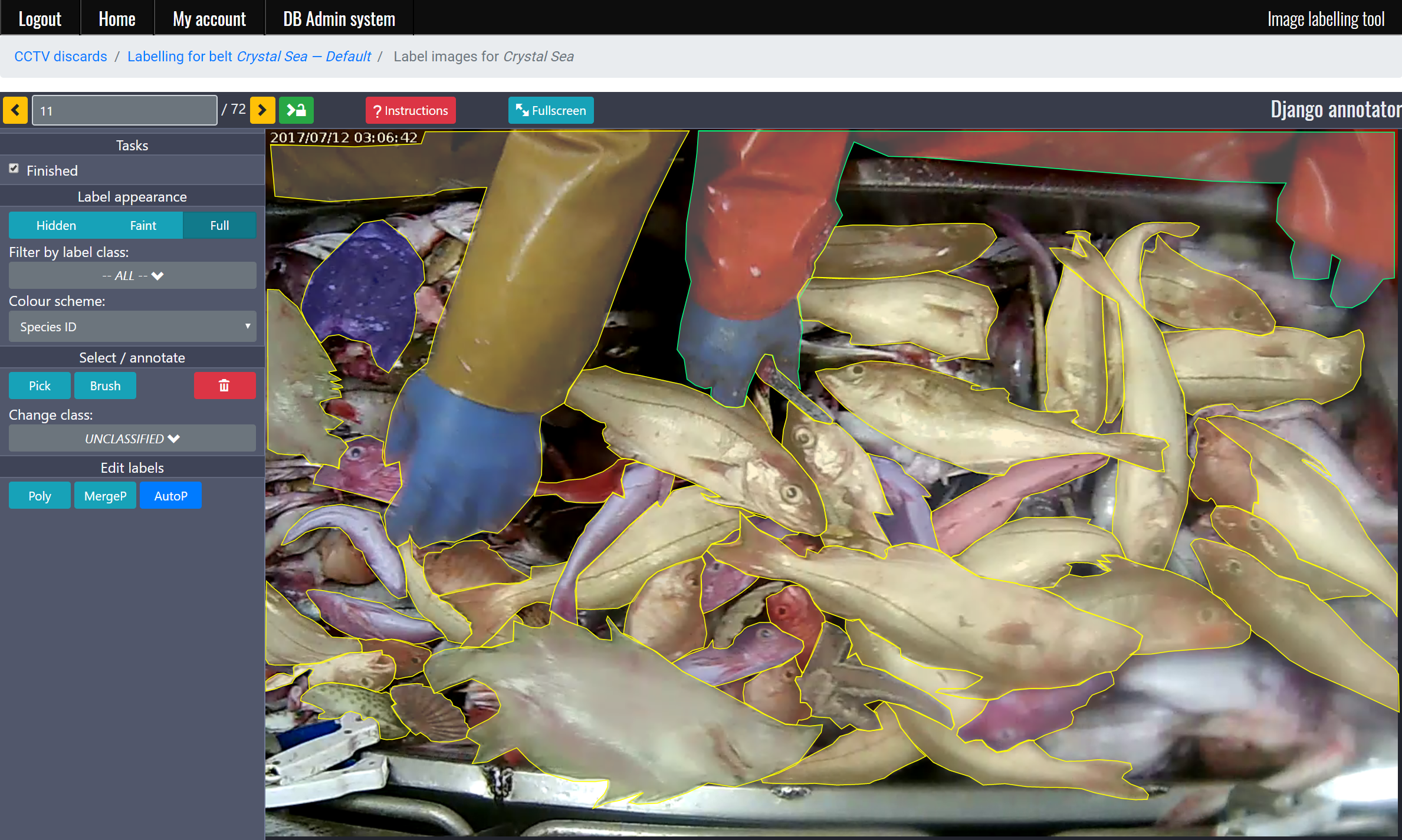}
  \caption{Web-based segmentation annotation tool} 
\label{fig:dataset:seg:annot_tool}
\end{figure}

While the aforementioned tools allow for a fast and effective work-flow, manually annotating fish is a labour intensive task, especially in dense images such as the one seen in Figure~\ref{fig:dataset:seg:annot_tool}.
To this end, we use an implementation\footnote{Available from \url{https://github.com/Britefury/dextr}} of the DEXTR~\cite{Maninis:DEXTR} to significantly speed up the outlining process.
The user clicks four extreme points around a fish, where the extreme points are on the top, bottom, left and right-most edges of the fish outline, after which DEXTR predict the mask for the fish, alleviating the need to draw the outline manually.
Manual outlines created using the annotation tool are used to fine-tune a DEXTR model trained on the Pascal VOC 2012~\cite{Everingham:PascalVOC2012} data set, resulting in a model that produces high quality outlines for fish.

When handling an image for which no hand-made annotations are available, we apply a Mask-RCNN~\cite{He:MaskRCNN} instance segmentation model trained on the annotated images in our data set to produce a set of initial annotations.
This combined with the use of DEXTR to fix any errors in the initial annotations allows us to quickly annotate images with minimal user effort.
The improved annotations are then used to train improved Mask-RCNN and DEXTR models, resulting in a cyclic workflow that improves the models iteratively.

\section{Species ID dataset summary}
\label{sec:spec_id_ds_summ}

The species ID dataset is summarised in Table~\ref{tab:species:commercial_data}.

\begin{table}[h!t]
  \begin{center}
  \scriptsize
  \begin{tabular}{|l|rrrrrr|r|}
  \hline
  Species                  & Vessel A  & Vessel B  & Vessel C  & Vessel D  & Vessel E  & Vessel F  & Total     \\
  \hline
  Cod                      & 123       & 146       & 99        & 73        & 0         & 6         & 447       \\
  Haddock                  & 308       & 459       & 713       & 1855      & 51        & 2112      & 5498      \\
  Whiting                  & 43        & 145       & 95        & 126       & 20        & 17        & 446       \\
  Saithe                   & 1002      & 11        & 368       & 96        & 0         & 0         & 1477      \\
  Hake                     & 283       & 32        & 198       & 48        & 3         & 14        & 578       \\
  Monk                     & 5         & 15        & 8         & 3         & 224       & 79        & 334       \\
  Mackerel                 & 16        & 0         & 15        & 1         & 0         & 0         & 32        \\
  Horse mackerel           & 33        & 0         & 48        & 19        & 0         & 0         & 100       \\
  Norway pout              & 102       & 44        & 24        & 60        & 6         & 3         & 239       \\
  Plaice                   & 4         & 142       & 18        & 1         & 361       & 8         & 534       \\
  Long rough dab           & 5         & 3         & 0         & 1         & 25        & 0         & 34        \\
  Common dab               & 0         & 10        & 7         & 1         & 28        & 0         & 46        \\
  Grey gurnard             & 146       & 18        & 29        & 36        & 58        & 443       & 730       \\
  Red gurnard              & 0         & 1         & 0         & 0         & 52        & 156       & 209       \\
  Gurnard (generic)        & 237       & 763       & 311       & 269       & 624       & 2897      & 5101      \\
  Dover sole               & 0         & 0         & 0         & 0         & 355       & 5         & 360       \\
  Lemon sole               & 3         & 19        & 0         & 2         & 131       & 7         & 162       \\
  Dog fish                 & 1         & 0         & 0         & 1         & 148       & 226       & 376       \\
  Sea bass                 & 0         & 0         & 0         & 0         & 1         & 0         & 1         \\
  John Dory                & 0         & 0         & 0         & 0         & 0         & 15        & 15        \\
  Megrim                   & 0         & 2         & 0         & 0         & 8         & 98        & 108       \\
  Ling                     & 12        & 2         & 3         & 1         & 0         & 0         & 18        \\
  Herring                  & 23        & 1         & 193       & 192       & 1         & 0         & 410       \\
  Bib                      & 0         & 0         & 0         & 0         & 411       & 17        & 428       \\
  Brill                    & 0         & 0         & 0         & 0         & 14        & 8         & 22        \\
  Turbot                   & 0         & 0         & 1         & 0         & 0         & 2         & 3         \\
  Boar fish                & 0         & 0         & 0         & 0         & 4         & 998       & 1002      \\
  Argentines (generic)     & 26        & 67        & 31        & 3         & 0         & 0         & 127       \\
  Witch                    & 0         & 1         & 0         & 0         & 0         & 1         & 2         \\
  Catfish                  & 0         & 1         & 1         & 0         & 0         & 0         & 2         \\
  Cuttlefish               & 0         & 0         & 0         & 0         & 3         & 1         & 4         \\
  Norway haddock           & 4         & 0         & 0         & 0         & 0         & 0         & 4         \\
  Red mullet               & 0         & 0         & 0         & 0         & 1         & 3         & 4         \\
  Conger eel               & 0         & 0         & 0         & 0         & 0         & 1         & 1         \\
  Common dragonet          & 0         & 0         & 0         & 0         & 138       & 65        & 203       \\
  Skate/ray                & 51        & 273       & 18        & 28        & 18        & 43        & 431       \\
  Squid                    & 0         & 0         & 0         & 0         & 0         & 74        & 74        \\
  Shark                    & 2         & 0         & 0         & 17        & 9         & 2         & 30        \\
  Brown Crab               & 0         & 0         & 0         & 0         & 0         & 0         & 0         \\
  European lobster         & 0         & 0         & 0         & 0         & 0         & 0         & 0         \\
  Crawfish                 & 0         & 0         & 0         & 0         & 0         & 0         & 0         \\
  Fish unidentifiable      & 763       & 894       & 838       & 2257      & 200       & 2815      & 7767      \\
  Flat (generic)           & 28        & 163       & 21        & 68        & 130       & 128       & 538       \\
  \hline
  Total                    & 3220      & 3212      & 3039      & 5158      & 3024      & 10244     & 27897     \\
  \hline
  
  
  \end{tabular}

  \caption[Summary of species ID dataset]{Summary of species identification dataset. Note that unsupervised samples are not included.}
  \label{tab:species:commercial_data}
  
  \end{center}
\end{table}

\clearpage

\section{Species ID confusion matrices}
\label{sec:spec_id_cm}

\begin{figure}
  \centering
  \includegraphics[width = 1.0\textwidth]{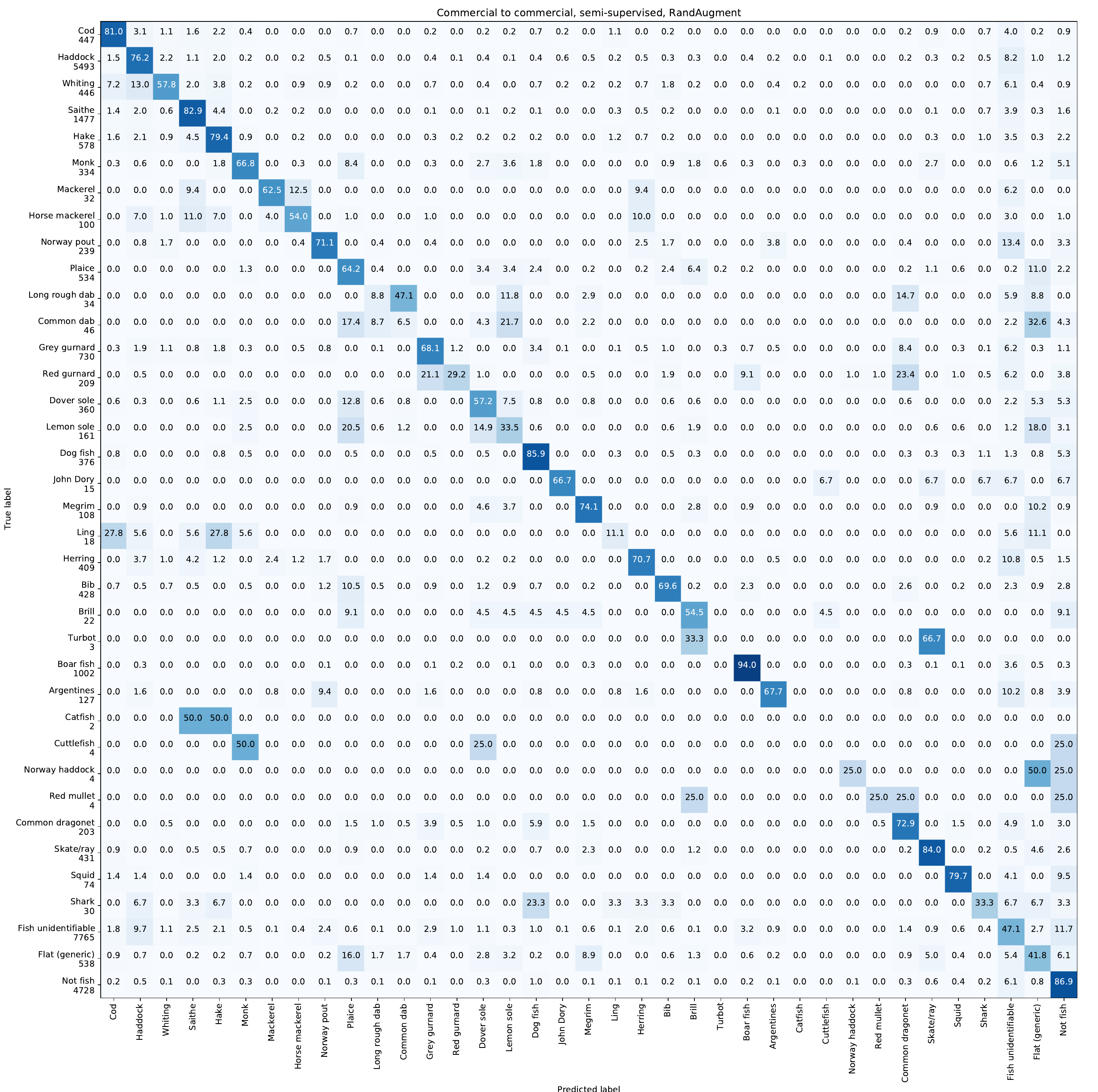}
  \caption[Species ID confusion matrix from semi-supervised training with RandAugment]{Confusion matrix for semi-supervised learning using Mean Teacher with RandAugment. Mean class accuracy is 53.8\%.
   Values along the left side below each class name indicate the number of samples in that class.}
\label{fig:species:cm_commercial_to_commercial_semuisup_ra}
\end{figure}

The performance of our classifier trained with semi-supervised Mean Teacher with RandAugment is illustrated by the confusion matrix shown in Figure~\ref{fig:species:cm_commercial_to_commercial_semuisup_ra}.
Briefly, each row of a confusion matrix shows how samples belonging to the ground truth class corresponding to the row were classified by the model.
The diagonal of the matrix shows the number of samples correctly classified, while off-diagonal elements show mis-classifications.

\begin{figure}
  \centering
  \includegraphics[width = 1.0\textwidth]{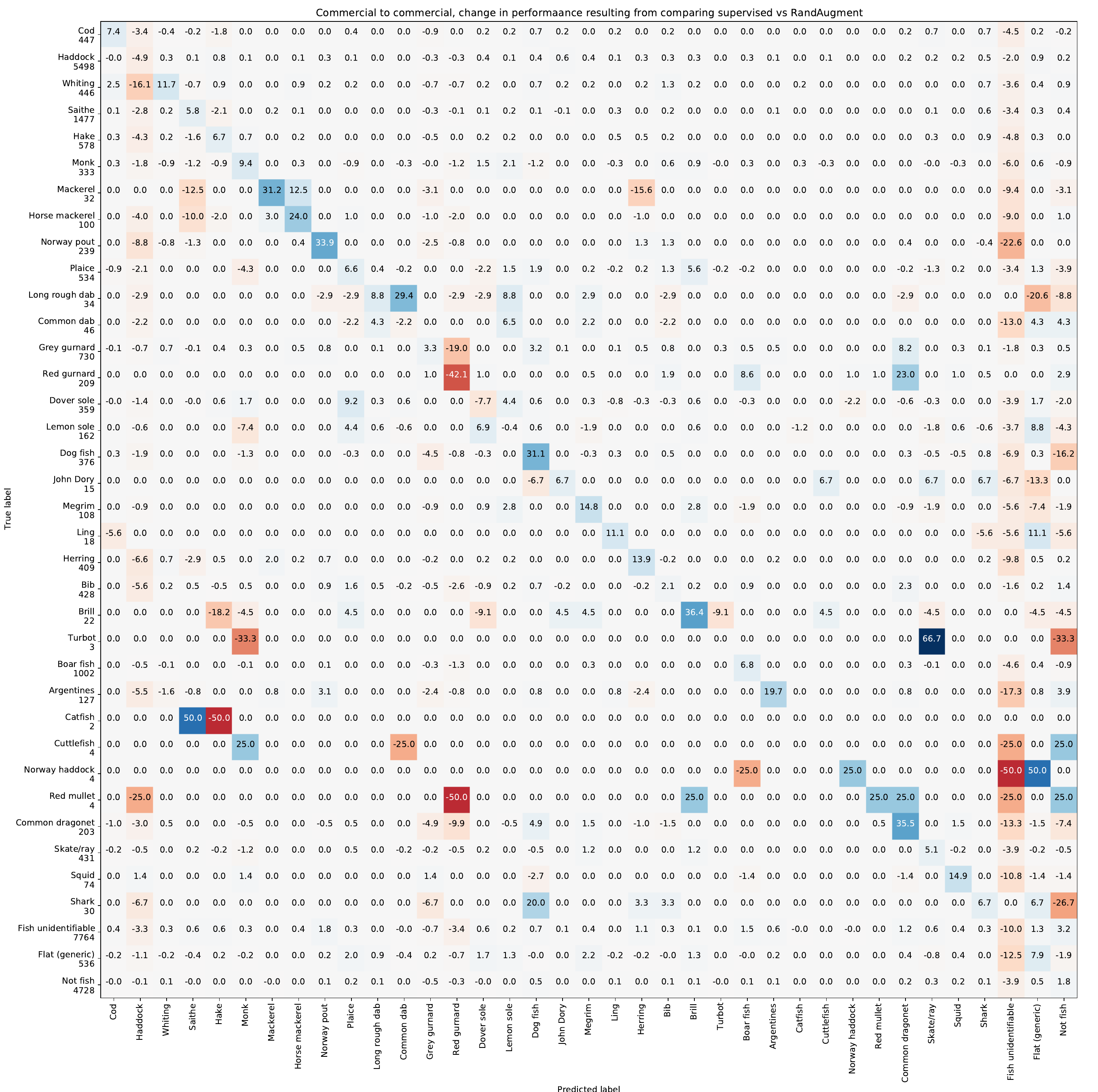}
  \caption[Species ID confusion matrix comparison of supervised and semi-supervised training with RandAugment]{Confusion matrix comparison of supervised and semi-supervised learning (with RandAugment) results.
  The plot represents the increase (blue) or decrease (red) in predictions resulting from applying semi-supervised learning.
  The values are the result of subtracting the supervised confusion matrix (not shown) from the semi-supervised confusion matrix (see Figure|~\ref{fig:species:cm_commercial_to_commercial_semuisup_ra}).
  Values along the left side below each class name indicate the number of samples in that class.
  }
\label{fig:species:cm_delta_commercial_to_commercial_sup_vs_semisup_ra}
\end{figure}

The improvement arising in comparison to plain supervised learning is illustrated in the confusion matrix comparison plot shown in Figure~\ref{fig:species:cm_delta_commercial_to_commercial_sup_vs_semisup_ra}.
We observe an improvement in performance indicated by the positive values -- indicating an increase -- running along the diagonal and the negative values (a decrease in prediction) outside the diagonal.
We note that off-diagonal positive values (resulting in a decrease in overall performance) tend to occur for classes with poor representation, e.g. long rough dab (34 samples), common dab (46), lemon sole(162), turbot (3), catfish (2), cuttlefish (4) and red mullet (4).
We note that the use of RandAugment increases the number of the relatively well represented grey (730) and red (209) gurnards mis-predicted as common dragonet (203).

\section{Discard species mapping}
\label{sec:discard_species_map}

Upon inspecting the ground truth discard counts we found that the set of classes used does not match those used during training.
As a consequence, a mapping between the classes had to be established.
Our class mappings from ground truth classes to training classes are given in Tables~\ref{tab:count:count_cls_map_mss} and \ref{tab:count:count_cls_map_cefas}.
Classes with identical names were trivially matched with one another.
Other classes were matched due to their similarity, \eg Atlantic herring (vessels A, B, C, \& D ground truth) with Herring (training).
Other less precise matches were made, but with a higher penalty.
The penalty value is used during our evaluation (described below) to match individual ground truth discards with predicted discards.

\begin{table}[h!]
  \begin{center}
  \scriptsize
  \begin{tabular}{lll}
  \toprule
  Vessel A, B, C \& D ground truth species &    Training species & Penalty \\
  \midrule

                            Anglerfish &                Monk &       1 \\
                           Argentines  &          Argentines &       1 \\
                      Atlantic herring &             Herring &       1 \\
               Atlantic Horse Mackerel &      Horse mackerel &       1 \\
                     Atlantic Mackerel &            Mackerel &       1 \\
                            Blue Skate &           Skate/ray &       1 \\
                          Blue Whiting &                  -- &         \\
                               Catfish &                  -- &         \\
                                   Cod &                 Cod &       1 \\
                            Common dab &          Common dab &       1 \\
                            Cuttlefish &          Cuttlefish &       1 \\
                     Common Squids nei &               Squid &       1 \\
          Dogfish Sharks nei / Dogfish &            Dog fish &       1 \\
                           Edible crab &          Brown Crab &       1 \\
                         European Hake &                Hake &       1 \\
                       European Plaice &              Plaice &       1 \\
   Flatfishes nei / Flatfish (generic) &      Flat (generic) &       1 \\
      Gurnards nei / Gurnard (generic) &        Grey gurnard &       1 \\
      Gurnards nei / Gurnard (generic) &         Red gurnard &       1 \\
                           Red gurnard &         Red gurnard &       1 \\
      Gurnards nei / Gurnard (generic) &   Gurnard (generic) &       1 \\
                               Haddock &             Haddock &       1 \\
                                  Hake &                Hake &       1 \\
                               Herring &             Herring &       1 \\
                        Horse Mackerel &      Horse Mackerel &       1 \\
                            Lemon Sole &          Lemon sole &       1 \\
                                  Ling &                Ling &       1 \\
                        Long rough dab &      Long rough dab &       1 \\
                              Mackerel &            Mackerel &       1 \\
                           Megrims nei &              Megrim &       1 \\
                              Monkfish &                Monk &       1 \\
                                Morays &                  -- &         \\
                        Norway Haddock &      Norway haddock &       1 \\
                        Norway Lobster &    European lobster &       2 \\
                           Norway Pout &         Norway pout &       1 \\
                        Norway Redfish &      Norway haddock &       1 \\
                               Octopus &                  -- &         \\
                                Plaice &              Plaice &       1 \\
                              Poor Cod &                  -- &         \\
                         Raja rays nei &           Skate/ray &       1 \\
 Rays and Skates nei / Rays and Skates &           Skate/ray &       1 \\
                        Red Stone Crab &          Brown Crab &       2 \\
                                Saithe &              Saithe &       1 \\
                                 Squid &               Squid &       1 \\
                           Tusked Goby &                  -- &         \\
             Tuskfishes nei / Tuskfish &                  -- &         \\
                               Whiting &             Whiting &       1 \\
                        Witch Flounder &               Witch &       1 \\
                        Unidentifiable & Fish unidentifiable &       1 \\

  \bottomrule
  \end{tabular}\caption[Class mapping for vessels A, B, C, \& D]{Species class mapping from ground truth classes used for vessels A, B, C and D to the set of classes used in the training set.
  Blank rows indicate that no training class was matched to the given ground truth class.
  Multiple ground truth species separated by slashes (/) indicate that slightly different species names were used by our expert analysts in the validation and test set ground truths.}
  \label{tab:count:count_cls_map_mss}
  \end{center}
\end{table}

\begin{table}[h!]
  \begin{center}
  \tiny
  \begin{tabular}{lll}
  \toprule
       Vessel E \& F ground truth species &    Training species & Penalty \\
  \midrule

                            Angler Fishes &                Monk &       1 \\
                         Atlantic Bobtail &               Squid &       1 \\
                             Atlantic Cod &                 Cod &       1 \\
                       Bib (Pout-Whiting) &                 Bib &       1 \\
                               Blonde Ray &           Skate/ray &       1 \\
                                Boar Fish &           Boar fish &       1 \\
                          Common Dragonet &     Common dragonet &       1 \\
             Cuttlefish (With Cuttlebone) &                 --  &         \\     
                               Cuckoo Ray &           Skate/ray &       1 \\
                                      Dab &          Common dab &       1 \\
                              Edible Crab &          Brown Crab &       1 \\
              Epibenthic Mix Unidentified & Fish unidentifiable &       1 \\
                    European Common Squid &               Squid &       1 \\
                      European Conger Eel &          Conger eel &       1 \\
                            European Hake &                Hake &       1 \\
                         European Lobster &    European lobster &       1 \\
                        European Mackerel &            Mackerel &       1 \\
                          European Plaice &              Plaice &       1 \\
                        Fish Without Jaws & Fish unidentifiable &       1 \\
                                 Flatfish &      Flat (generic) &       1 \\
                                 Flounder &      Flat (generic) &       2 \\
                            Great Scallop &                  -- &         \\
                      Greater Spider Crab &          Brown Crab &       2 \\
                           Gurnards Indet &   Gurnard (generic) &       1 \\
                           Gurnards Indet &        Grey gurnard &       1 \\
                           Gurnards Indet &         Red gurnard &       1 \\
                                  Haddock &             Haddock &       1 \\
                                  Herring &             Herring &       1 \\
Horse-Mackerel (Scad) / Scad (Horse Mackerel) Indet & Horse mackerel &  1 \\
                                John Dory &           John Dory &       1 \\
                               Lemon Sole &          Lemon sole &       1 \\
                   Lesser Spotted Dogfish &            Dog fish &       1 \\
                       Lesser Weever Fish &                  -- &         \\
                               Loligo Spp &               Squid &       1 \\
         Long-Rough Dab (American Plaice) &      Long rough dab &       1 \\
                               Lumpsucker &                  -- &         \\
                                   Megrim &              Megrim &       1 \\
                              Norway Pout &         Norway pout &       1 \\
    Nurse Hound (Greater Spotted Dogfish) &            Dog fish &       1 \\
                            Octopus Indet &                  -- &         \\
                                 Poor Cod &                  -- &         \\
                            Queen Scallop &                  -- &         \\
                                Sand Sole &                  -- &         \\
                                Sandy Ray &                  -- &         \\
                                Scaldfish &                  -- &         \\
            Skate Indet / Skates and Rays &           Skate/ray &       1 \\
             Small-eyed Ray (Painted Ray) &           Skate/ray &       1 \\
                        Sole (Dover Sole) &          Dover sole &       1 \\
                               Sole Indet &                  -- &         \\
                                Solenette &                  -- &         \\
                      Spiny Lobster Indet &    European lobster &       2 \\
                              Spotted Ray &           Skate/ray &       1 \\
                        Squids and Octopi &               Squid &       1 \\
                      Starry Smooth Hound &           Dog fish  &       1 \\
                       Striped Red Mullet &          Red mullet &       1 \\
                           Thickback Sole &      Flat (generic) &       2 \\
                            Thornback Ray &           Skate/ray &       1 \\
                   Undulate (Painted) Ray &           Skate/ray &       1 \\
                                  Whiting &             Whiting &       1 \\
                                 Flatfish &      Flat (generic) &       1 \\
                           Unidentifiable & Fish unidentifiable &       1 \\
  
  \bottomrule
  
  \end{tabular}\caption[Class mapping for vessels E and F]{Species class mapping from ground truth classes used for vessels E and Fs to the set of classes used in the training set.
  }
  \label{tab:count:count_cls_map_cefas}
  \end{center}
\end{table}

\end{document}